\documentclass[conference]{IEEEtran}

\usepackage[latin1, utf8]{inputenc}
\usepackage{tikz}
\usepackage{fancyhdr}

\IEEEoverridecommandlockouts
\usepackage{pdfpages}
\usepackage{todonotes}
\usepackage{cite}
\usepackage{subfig}
\usepackage{amsmath,amssymb,amsfonts}
\usepackage{algorithmic}
\usepackage[ruled,linesnumbered]{algorithm2e}
\usepackage{graphicx}
\usepackage{textcomp}
\usepackage{multirow}
\usepackage{tablefootnote}
\usepackage{threeparttable}
\usepackage{blindtext}
\usepackage[inline]{enumitem}
\usepackage{xcolor}
\usepackage{layouts}
\usepackage{svg}

\bstctlcite{IEEEexample:BSTcontrol}
\usepackage{xcolor, bm}

\usepackage{gensymb}

\DeclareMathOperator*{\argmin}{arg\,min}

\begin{document}

\bstctlcite{IEEEexample:BSTcontrol}
\title{
One-Shot Online Testing of Deep Neural Networks Based on Distribution Shift Detection
}

\author{\IEEEauthorblockN{Soyed Tuhin Ahmed, Mehdi B. Tahoori}
\IEEEauthorblockA{
\textit{Department of Computer Science},
\textit{Karlsruhe Institute of Technology (KIT)}\\
soyed.ahmed@kit.edu, mehdi.tahoori@kit.edu}
}

\maketitle



\begin{abstract}
Neural networks (NNs) are capable of learning complex patterns and relationships in data to make predictions with high accuracy, making them useful for various tasks. However, NNs are both computation-intensive and memory-intensive methods, making them challenging for edge applications. To accelerate the most common operations (matrix-vector multiplication) in NNs, hardware accelerator architectures such as computation-in-memory (CiM) with non-volatile memristive crossbars are utilized. Although they offer benefits such as power efficiency, parallelism, and nonvolatility, they suffer from various faults and variations, both during manufacturing and lifetime operations. This can lead to faulty computations and, in turn, degradation of post-mapping inference accuracy, which is unacceptable for many applications, including safety-critical applications. Therefore, proper testing of NN hardware accelerators is required. In this paper, we propose a \emph{one-shot} testing approach that can test NNs accelerated on memristive crossbars with only one test vector, making it very suitable for online testing applications. Our approach can consistently achieve $100\%$ fault coverage across several large topologies with up to $201$ layers and challenging tasks like semantic segmentation. Nevertheless, compared to existing methods, the fault coverage is improved by up to $24\%$, the memory overhead is only $0.0123$ MB, a reduction of up to $19980\times$ and the number of test vectors is reduced by $10000\times$. 

\end{abstract}


\section{Introduction}\label{sec:introduction}

Deep learning algorithms have been the driving force behind substantial advancements in various domains, such as computer vision, natural language processing, and speech recognition. Recently, deep learning algorithms have been increasingly deployed in safety- and security-critical domains such as autonomous driving, medical imaging, and malware detection. At the heart of deep learning systems are multi-layered neural networks (NNs) that learn hierarchical representations from the training dataset and make actionable predictions on inference data. Despite the algorithmic success of NNs, they are computationally demanding, and their conventional hardware implementation suffers from a memory bottleneck due to von Neumann architectures, where memory and processing units are physically separated, leading to significant data movement and energy consumption.

Therefore, several specialized architectures and hardware accelerators, such as computation-in-memory (CiM) architectures~\cite{yu2018neuro}, have been explored to accelerate NNs in hardware. CiM leverages emerging non-volatile memory (NVM) technologies, such as Resistive Random-Access Memory (ReRAM)~\cite{akinaga2010resistive}, Phase Change Memory (PCM)~\cite{wong2010phase}, and Spin Transfer Torque Magnetic Random Access Memory (STT-MRAM)~\cite{apalkov2013spin}, to perform computations directly in memory, mitigating the memory bottleneck. Emerging NVM technologies offer benefits such as zero leakage power, non-volatility, high switching speed, and endurance compared to conventional CMOS-based memories.

However, emerging NVM technologies exhibit several post-manufacturing and online non-idealities, including read disturb error~\cite{7035342}, retention faults~\cite{hofmann2014comprehensive}, manufacturing  variations, and online thermal variations~\cite{niu2010impact}. These non-idealities can adversely impact the online functionality of memristive chips for deep learning applications and negatively impacts the prediction ability of the NNs~\cite{ahmed2022process}. Therefore, testing such hardware systems is crucial to ensuring their reliability and correct functionality, especially in safety-critical applications.

\begin{figure}
    \centering
    \includegraphics[width=\linewidth]{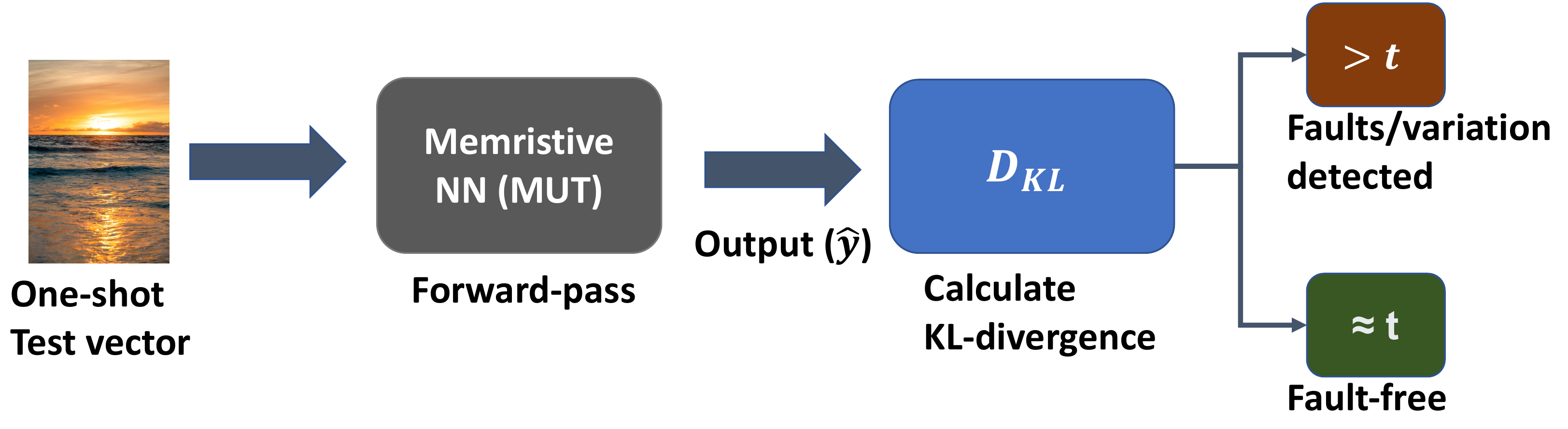}
    \caption{Flow diagram of our proposed one-shot testing approach. A KL-divergence value greater than a predefined threshold indicates faults or variation in the memristive NN.}
    \label{fig:flow}
\end{figure}

Nevertheless, testing NN hardware accelerators present a unique set of challenges due to the complexity, inherent non-linearity, and vast number of layers and parameters. Unlike traditional hardware or software testing approaches, NNs cannot be exhaustively tested with all possible input combinations, as they can be millions, leading to high testing overhead. Furthermore, specialized CiM-based NN hardware accelerators do not contain conventional digital Design for Test (DfT) infrastructure, such as scan chains. A testing approach that does not require access to training data, treats the NN as an intellectual property (IP), that is it does not require access to the intermediate results or backdoors to the model (non-invasive) but can test the NN in a \emph{one-shot} is desired. 

One-shot testing, which is the extreme form of test compaction, can test an NN model and its hardware realization with a single test vector and forward pass. It can minimize testing time, the computation required, and system downtime to a minimum. Even memristor chip-specific testing approaches can lead to long system downtime due to the large number of test vectors~\cite{ahmedETS23}. Longer system downtime can be unacceptable for many applications, including "always-on" scenarios, e.g., real-time object detection and tracking, voice assistants, anomaly detection, and predictive maintenance, particularly in mission-critical applications.

In this paper, we propose a comprehensive \emph{one-shot} testing framework for CiM-based memristive deep learning hardware accelerators that treat NNs as a black box and do not require access to training datasets or intermediate results. Our approach is capable of testing large-scale NNs with hundreds of layers using a single testing vector, significantly reducing the number of forward passes and computational overhead during testing. We evaluate our approach on several
large CNN topologies with up to $201$ layers and several difficult tasks, e.g., ImageNet classification with $1000$ classes, and semantic segmentation on real-world biomedical segmentation. Nevertheless, we were able to consistently achieve $100\%$ test coverage of different fault types and fault severeness.

The rest of the paper is organized as follows: Section~\ref{sec:background} provides background information on deep learning, NVM technologies, and their non-idealities. Section~\ref{sec:propsoed} describes our proposed one-shot testing method in detail, Section~\ref{sec:result} describes the fault injection framework, evaluates our approach, and presents the results, and finally, Section~\ref{sec:conclusion} concludes the paper.
\section{Preliminaries}\label{sec:background}

\subsection{Memristor Devices and Non-idealities}
\begin{table}
\caption{Notations used in this paper.}
\resizebox{\linewidth}{!}{
\begin{tabular}{|@{}c@{}|c@{}|@{}c@{}|c@{}|}
\hline
Symbol & Explanation                   & Symbol & Explanation                            \\ \hline
   $t$    & Threshold for fault detection & $\hat{\mathcal{N}}$      & Output distribution of a memristive NN \\ \hline
   $D_{\mathrm{KL}}(\hat{\mathcal{N}} \parallel \mathcal{N})$    & KL divergence between $\hat{\mathcal{N}}$ and $\mathcal{N}$ &     $\mathcal{N}$ & Expected output distribution (unit Gaussian)           \\ \hline
$\mathbf{W}$      & Weight matrix                 & $\Bar{x}$      & One-shot test vector                   \\ \hline
$\mathbf{b}$      & Bias vector                   & $\mathcal{L}$      & Loss function                          \\ \hline
$\hat{y}$      & Output of the Memristive NN   & $\mathcal{F}$      & Pre-trained NN                         \\ \hline
$Q$      & Learning rate decay rate      &        $\alpha$ & Learning rate                          \\ \hline
$\nabla \mathcal{L}$       & Gradient from backpropagation &   $\eta_0$     & Noise scale for variations             \\ \hline
   $\mathcal{P}_{flip}$     & Percentage of faults          &        $y'$ & The ground truth for optimization      \\ \hline
   $\mathbf{z}$     & Intermediate activations          &        $\theta$ & Learable parameters of a NN      \\ \hline
   $\mu$     & Mean          &        $\sigma^2$ & Variance      \\ \hline
   $N$     & Number  of output classes          &        $\mathcal{M}$ & Number of Monte Carlo faults runs      \\ \hline
\end{tabular}
}
\end{table}

Memristor technology including ReRAM~\cite{akinaga2010resistive}, PCM~\cite{wong2010phase}, and  STT-MRAM~\cite{apalkov2013spin} are two-terminal nanoscale devices that are the basic building block of in-memory computing for the acceleration of NNs. The number of stable states varies from technology, e.g., STT-MRAM can be programmed to Low Resistance State $LRS$/High Conductance State $G_{off}$ or High Resistance State $HRS$/Low Conductance State $G_{on}$ but ReRAM and PCM can be programmed into multiple stable states~\cite{aziza2021multi}. Multilevel cells can also be designed using multiple STT-MRAM devices.

Despite their promising characteristics, due to a number of factors, memristive devices exhibit a number of non-idealities~\cite{nair2020defect, chen2014rram, le2020overview, nair2017vaet, bishnoi2014read, zhou2020noisy, kim2020efficient} that can be broadly categorized as either permanent or soft faults.

Permanent faults refer to those that irreversibly alter the conductance state of memristor cells, preventing them from being programmed to the desired resistance/conductance state for encoding NN parameters. Cells with permanent faults cannot be restored to their original fault-free values.

On the contrary, soft faults refer to those that temporarily alter the conductance state of memristor cells but can still cause deviations in NN parameters. Faulty memristor cells can, however, be restored to their original values.

Irrespective of the specific type of fault, they occur during both the manufacturing process and in-field operation. As a result, the model parameters and activations of NNs can deviate from their expected values  after their hardware mapping (post-mapping) due to manufacturing faults and post-deployment due to runtime faults. In this section, common memristor device faults and their corresponding fault models are discussed.

\paragraph{Stuck-at faults} Among all kinds of hard faults, stuck-at faults  appear more frequently in memristive crossbars. Suck-at faults are modelled as the memristor cell conductance can become stuck at high conductance (stuck-at-$G_{on}$) or low conductance (stuck-at-$G_{off}$). Depending on the occurrence factor, stuck-at faults can be categorized as either soft faults or permanent faults. Stuck-at faults caused by limited endurance from repeated reading are categorized as soft faults, whereas the manufacturing defects that cause stuck-at faults are categorized as permanent faults. In a memristive crossbar array, stuck-at faults are randomly distributed and can be as high as 10\%~\cite{chen2014rram}. 
Defects like stuck-open or short can also be modeled as stuck-at-$G_{off}$ and stuck-at-$G_{on}$\cite{nair2020defect, chen2014rram}. Consequently, the parameters of the memristive NN implementation deviate from their initial values. The parameter bit change, depending on encoding, can be represented as either stuck-at-0 or stuck-at-1.

\paragraph{Manufacturing and In-field Variations} Device variability occurs when the conductance of memristors exhibits distribution rather than a fixed value due to factors such as manufacturing process variations. In-field variations, on the other hand, emerge from the dynamic changes in the memristor's environment, including temperature or other environmental factors, which can fluctuate the conductance. The effect of both types of variations can cause variations in the current sum in the bit-line in the crossbar and a reduction in the sensing margin, leading to incorrect sensed values. 

\paragraph{Read/write disturbance} Memristor reading (inference) and writing (parameter mapping) can both be affected by read and write currents impacting other memristor cells sharing the same bit-line in the crossbar array. Such faults can lead to unintentional switching of memristor conductance states during read operations. Moreover, write disturbance faults influence the data (NN parameters) stored in memristor cells\cite{chen2014rram, bishnoi2014read}. 

\paragraph{Slow-Write Fault} During NN parameter mapping, defective memristor cells might experience longer write delays, referred to as slow-write faults. In ReRAM, slow-write faults can emerge from repeated write operations. Switching in MTJ and PCM is inherently stochastic, causing non-deterministic write delays even when the environmental factors remain constant. A write failure can happen if the MTJ does not switch within a specified time or the switching pulse is truncated before the switching operation is completed~\cite{nair2017vaet, le2020overview}.  

\subsection{Neural Networks (NNs)}


Neural Networks (NNs) are computational models inspired by the structure and operation of biological neural networks. NNs comprised of multiple layers of neurons organized into a single input, single output, and multiple hidden layers. The input layer does not perform any computation, but only receives the input data. However, the hidden layers compute intermediate activations  $\mathbf{z}$, and the output layer generates the final results $\mathbf{y}$. The basic computation of a layer $l$ consists of the weighted sum of inputs $\mathbf{x}$ and the element-wise addition of bias. Afterwards, a non-linear activation function $\phi(\cdot)$ is applied. The overall mathematical computation of the NNs is as follows:

\begin{align} \label{eq:nn_forwardpass}
    \mathbf{z}^{(0)} &= \mathbf{x}, \\
    \mathbf{z}^{(l)} &= \phi^{(l)} \left( \mathbf{W}^{(l)} \mathbf{h}^{(l-1)} + \mathbf{b}^{(l)} \right), \quad l = 1, 2, \dots, L-1, \\
    \mathbf{\hat{y}} &= \Bar{\phi}^{(L)} \left( \mathbf{W}^{(L)} \mathbf{h}^{(L-1)} + \mathbf{b}^{(L)} \right),
\end{align}

where, $\mathbf{W}$, $L$, and  $\Bar{\phi}$ represent the weight matrix, the total number of layers, and the final transformation, e.g., SoftMax, respectively. SoftMax rescales the output values between $0$ and $1$.


NNs can be categorized based on their layer types and arrangements within the network. One popular type is the Convolutional Neural Network (CNN), which incorporates convolutional and linear layers. Another type is the Multi-layer Perceptron (MLP), which solely uses linear layers. CNNs are particularly powerful and are commonly applied in tasks that involve image, audio, and video. Thus, our methodology is assessed using state-of-the-art (SOTA) CNN architectures.

Normalization technique, such as batch normalization, is increasingly utilized to enhance the convergence speed and stability of the learning process. Batch normalization normalized the activations of each neuron during training prior to the application of two learnable parameters $\beta$ and $\gamma$ that scale and adjust the normalized activations as follows: 

\begin{equation}\label{eq:BN_eq}
    \mathbf{\Bar{z}}^{(l)} = \frac{\mathbf{W}^{(l)} \mathbf{x}^{(l-1)} + \mathbf{b}^{(l)} - \mathbf{\mu}^{(l)}}{\sqrt{\mathbf{\sigma}^{2(l)} + \epsilon}} \beta + \gamma,
\end{equation}

where, the batch mean and variance are denoted by $\mathbf{\mu}^{(l)}$ and $\mathbf{\sigma}^{2(l)}$, respectively. Also, $\epsilon$ is a small constant added for stability. 

The NN training procedure consists of learning the parameters $\theta$, which summarize all the learnable parameters given the training dataset $\mathcal{D} \subset (\mathbf{x}, \mathbf{y})$ with $N$ training examples by minimizing a task-specific loss function $\mathcal{L}$:

\begin{equation} \label{eq:nn_training_supervised}
\argmin{\bm{\theta}} = \frac{1}{N}\sum_{i=1}^N\mathcal{L}_{\theta}(\mathbf{y}_i, \mathbf{\hat{y}}_i).
\end{equation}

\subsection{Deep Learning Acceleration with Memristor-based Crossbars}

\begin{figure}
    \centering
    \includegraphics[width=\linewidth]{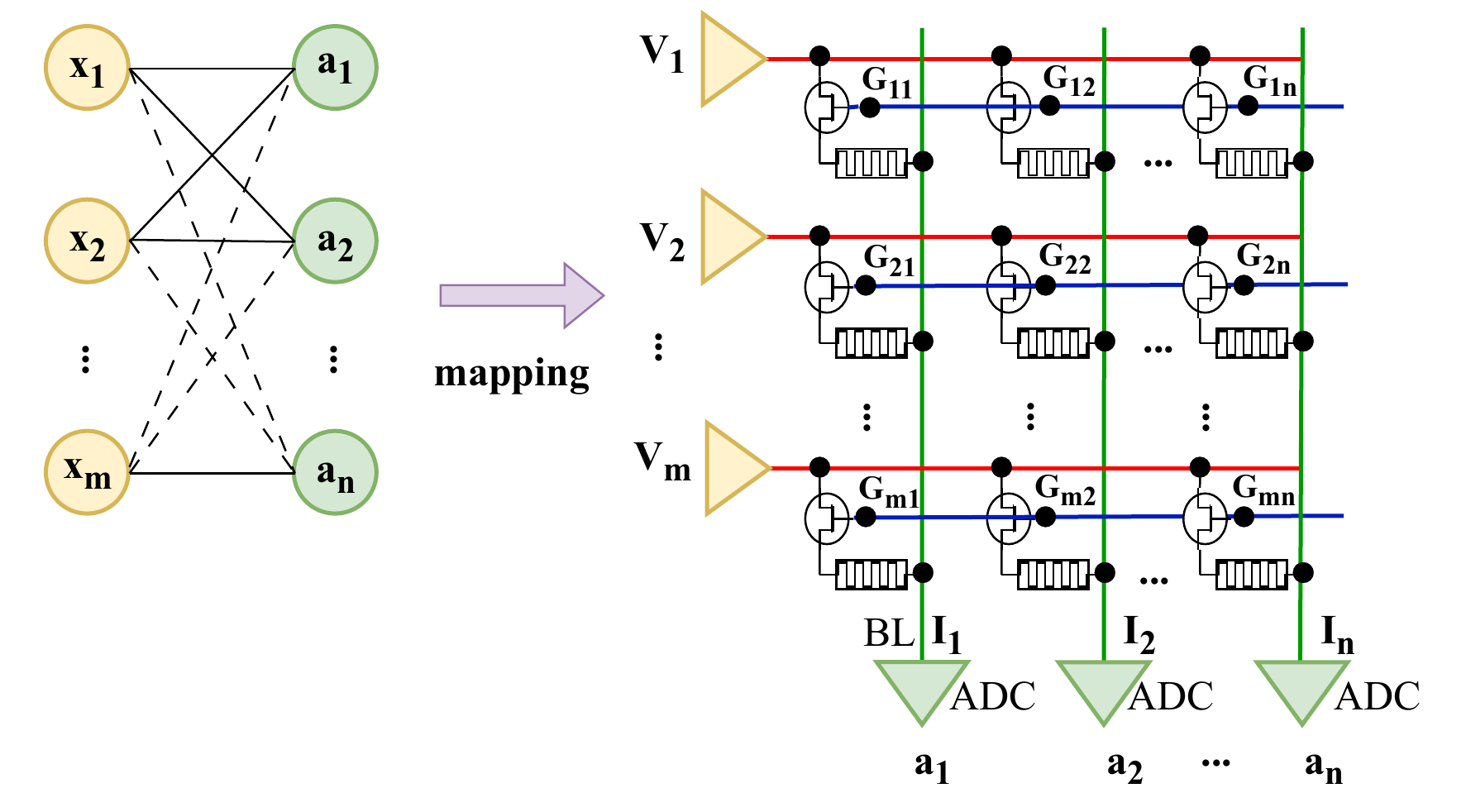}
    \caption{Mapping a layer of NN into a memristor-based crossbar array. }
    \label{fig:memristor_map}
\end{figure}

Memristive devices can be arranged into crossbar arrays, with each cross point consisting of a memristive device, as depicted in Fig.~\ref{fig:memristor_map}. Therefore, the weighted sum computation required for the inference stage of the NN can be carried out directly in the memory by leveraging Ohm's Law (V = IR) and Kirchhoff's Current Law at a constant $O(1)$ time without any data movement between the processing element and the memory.

Due to the finite number of conductance states of memristors, the trained parameters $\theta$ are first quantized to signed 8-bit precision using a post-training quantization approach, with negligible performance penalties. However, quantization-aware training should be performed for lower-bit precision quantization. 

Afterwards, the quantized parameters $\theta$ of the NN are mapped to the memristor-based crossbar arrays with NVM technology-specific encoding. For emerging NVM technology with two stable states, e.g., STT-MRAM, each bit of a parameter is encoded as high $G_{off}$ or low conductance $G_{on}$. Therefore, each cell in the crossbar array represents a single-bit ($0$ or $1$). For multilevel NVM technology, e.g., ReRAM or PCM with 128 resistance states, the sign can be represented with a single bit, $G_{off}$ (0) for positive, $G_{on}$ (1) for negative and the magnitude (0 to 127) can be represented with multi-level cells. A look-up table can be used to map the magnitudes to the conductance values of the ReRAM cells. 

The input vector $x$ is converted to continuous voltages and then streamed into the word-lines of the crossbar array for inference. Multiple word-lines of the crossbar are activated simultaneously for parallel computation, and the current that flows into the bit-line of the crossbar represents the result of the weighted sum operation. Ultimately, an Analog-to-Digital Converter (ADC) circuit digitizes the sensed currents. After that, other computations in the digital domain, e.g., bias addition, batch normalization, and non-linear activation operations are undertaken. Note that, ADC circuits are also subject to variations. However, we consider them to be robust to variations and do not add any noise to the NN activation.

Note that popular CNN topologies use skip connections, which allow information to bypass a few intermediate layers and add to the output of another layer. It can be implemented in a memristive crossbar NN by routing output signals through specific crossbars and adding the resulting outputs using digital summing circuits. However, since the NN computations are done sequentially, signals for the skip connections can be stored in the buffer memory.

\subsection{Related Works}
In the literature, several testing approaches have been proposed for testing memristor-based crossbar arrays. March-based algorithms serially program and read the memristor cells under test to a specified conductance level to identify faults~\cite{liu2016efficient}. However, March-based algorithms are not practical for memristor-mapped NN applications due to their large number of memory cells, which results in extended test times. Additionally, testing multi-level cells requires setting the memristive cells to all possible levels, further increasing test time.

An alternative approach for fault detection involves analyzing the deviation in the inference accuracy of either original training data or synthetic testing data in the presence of faults \cite{chen2021line, li2019rramedy, luo2019functional}. Synthetic testing data can be generated using adversarial examples \cite{li2019rramedy}, watermarking the training data, and re-training the NN on the testing data to create a backdoor \cite{chen2021line}. Although such methods efficiently detect deviations, they necessitate a large amount of testing data, on-chip storage (depending on the availability of on-chip retraining data), and an invasive test generation process. The performance of backdooring when common data augmentation techniques, such as corner padding and center-cropping, are used is unclear, since data augmentation can either partially or completely remove the watermarks. Additionally, watermarking relies on the translation invariance feature of CNN to achieve high accuracy on the test dataset and similar performance on the original task. However, since MLPs are not translation invariant, this method may not be suitable for MLPs. The work in \cite{luo2019functional} proposed back-propagating to the input image and using the gradient of the input image as standalone testing data or combining it with training data as a perturbation, similar to \cite{li2019rramedy} which employs the fast gradient sign method (FGSM).  However, their "pause-and-test" method leads to long periods of system downtime. The work presented in \cite{9693118} proposes monitoring the dynamic power consumption of crossbar arrays to detect faults. To achieve this, an adder tree is implemented to continuously monitor the dynamic power consumption, which adds hardware overhead.

A compact functional testing method has been studied in the work by~\cite{ahmed2022compact}. Their method can achieve high testing coverage with a sufficiently large number of testing vectors, typically ranging from 16 to 64. However, it has been observed that their method does not work well when the number of testing vectors is small, i.e., less than 10. Furthermore, their method relies on access to the training dataset.

In contrast, our one-shot vector generation and testing method
\begin{enumerate*}[label={\alph*)},font={\color{black!50!black}\bfseries}]
\item do not require access to training data, 
\item is non-invasive,
\item generalizable across different classes of NN,
\item requires only single testing queries,
\item needs negligible storage, power, and testing time, and 
\item can achieve high fault coverage.
\end{enumerate*}
\section{One-shot Testing of Memristive NNs} \label{sec:propsoed}

\subsection{Motivation and hypothesis of our approach}
\begin{figure*}
 \centering
 \includegraphics[width=0.99\linewidth]{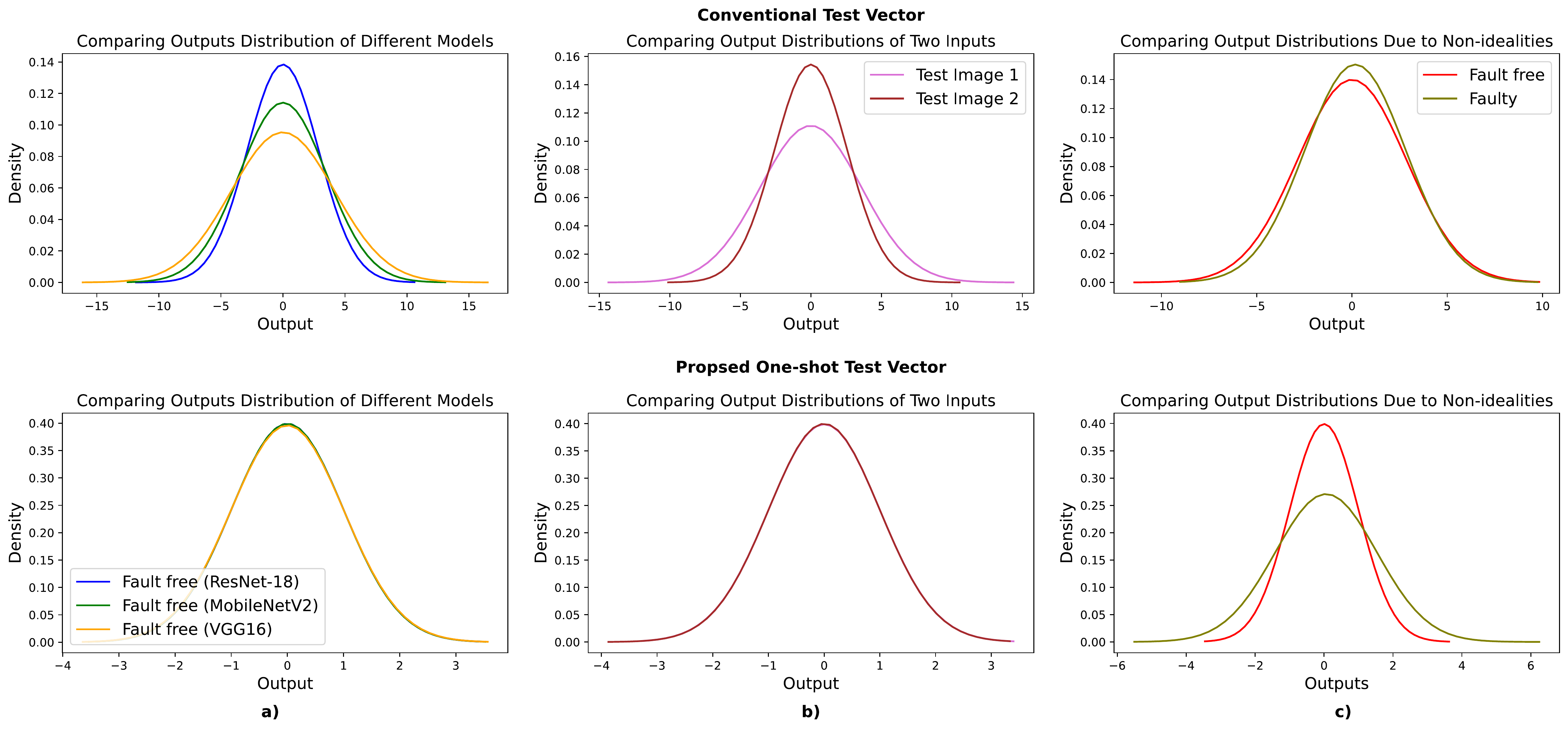}
 \caption{
 a) Change in output distribution depicted for different NN models but on the same test vector, b) comparison of the change in the output distribution for two different test vectors but on the same NN model (ResNet-18). While the conventional method reveals a change in output distribution across different models and test vectors, our approach ensures standardized output distributions (distributions overlap) irrespective of models or test vectors. c) We compared the relative change in output distribution for the same noise level between the proposed and conventional test vectors. The output distribution is more sensitive to noise for our proposed one-shot test vector. In the conventional method, the test vectors are randomly sampled from the ImageNet validation dataset.
 }
\label{fig:distribution_change}
 \end{figure*}

As discussed earlier, in the presence of faults or variations in the parameters of memristor-mapped NNs, their representation changes from the expected (trained) parameters, resulting in degradation in their performance. According to Equation~\ref{eq:nn_forwardpass}, non-ideal parameters will directly affect the weighted sum and in turn activation of a layer. Since the activation of a layer becomes the input to the following layer, the cascading effect of non-ideal parameters is likely to ultimately flow to the overall output $\hat{y}$  of the NNs. Therefore, the distribution $\hat{y}$ is also expected to change, as shown in Fig.~\ref{fig:distribution_change}(c). 

We introduce a novel \emph{one-shot testing} method based on the observation and hypothesis that faults and variations in memristive NN parameters influence the distribution of $\hat{y}$. Our approach aims to detect distribution shifts in the model output using a single test vector that is specifically designed to produce two distinct output distributions for faulty and fault-free cases. Consequently, faults and variations can be easily detected by evaluating the output distribution of a memristive NN after applying the one-shot test vector.

However, there are several challenges associated with this approach. The primary challenges include standardizing the output distribution using one test vector, estimating the change in distribution for pre-trained models, and designing an effective one-shot test vector for various model architectures. We discuss them in the following section with their respective solutions.

\subsection{Proposed deviation detection method}

Since the expected distribution of a model is unknown and likely varies from one model to another and from one test vector to another (as shown in the top half of Fig~\ref{fig:distribution_change}(a) and (b)), it is difficult to estimate the change in distribution for a pre-trained model. Therefore, we propose standardizing the output distribution of each \emph{model under test} (MUT) to a unit Gaussian distribution, $\hat{y} \sim \mathcal{N}(\mu\approx 0,\,\sigma^{2}\approx 1)\,$, which means zero mean $\mu\approx0$ and unit variance $\sigma^2\approx1$. 
Standardizing the output distribution is crucial for the one-shot testing method, as it ensures a consistent and comparable metric across various models and test vectors. Also, it reduces the likelihood of false-positive deviation detection and enhances the sensitivity to non-ideal parameters.

Let $\hat{y} \sim \hat{\mathcal{N}}(\mu,\,\sigma^{2})\,$ be the output distribution of a memristive NN model. Faults and variations in the parameters of memristive NNs can be detected by evaluating the Kullback–Leibler (KL) divergence between the expected output distribution $\mathcal{N}$ and the output distribution of memristive NNs as:

\begin{equation}\label{eq:loss_KL}
  D_{\mathrm{KL}}(\hat{\mathcal{N}} \parallel \mathcal{N}) = \sum_{i=1}^n \hat{\mathcal{N}}(i) \log\frac{\hat{\mathcal{N}}(i)}{\mathcal{N}(i)}  
\end{equation}

 which can be simplified for two normal distributions as:

\begin{equation}
    D_{\mathrm{KL}}(\hat{\mathcal{N}} \parallel \mathcal{N}) = \log\frac{\sigma_{\mathcal{N}}}{\sigma_{\hat{\mathcal{N}}}} + \frac{\sigma_{\hat{\mathcal{N}}}^2 + (\mu_{\hat{\mathcal{N}}} - \mu_{\mathcal{N}})^2}{2\sigma_{\mathcal{N}}^2} - \frac{1}{2}.
\end{equation}

We assumed that the distributions $\hat{\mathcal{N}}$ and $\mathcal{N}$ were discrete, since we quantized the parameters of the memristive NN. Here, we denote the mean and standard deviation of the output distribution $\hat{\mathcal{N}}$ of the memristive NN as $\mu_{\hat{\mathcal{N}}}$ and $\sigma_{\hat{\mathcal{N}}}$, respectively. Similarly, $\mu_{\mathcal{N}}$ and $\sigma_{\mathcal{N}}$ represent the mean and standard deviation of the expected output distribution $\mathcal{N}$. Since $\mu_{\mathcal{N}}$ and $\sigma_{\mathcal{N}}$ are defined as 0 and 1, respectively, the equation can be further simplified as:

\begin{equation}
    D_{\mathrm{KL}}(\hat{\mathcal{N}} \parallel \mathcal{N}) = \log\frac{1}{\sigma_{\hat{\mathcal{N}}}} + \frac{\sigma_{\hat{\mathcal{N}}}^2 + \mu_{\hat{\mathcal{N}}}^2}{2} - \frac{1}{2}.
\end{equation}

The KL divergence measures how one probability distribution differs from another. A larger value of $D_{\mathrm{KL}}(\hat{\mathcal{N}}\parallel \mathcal{N})$ indicates that the output distribution of the memristive NN is different from the expected distribution due to non-idealities in the parameters. Specifically, a threshold $t$ can be defined, where $D_{\mathrm{KL}}(\hat{\mathcal{N}}\parallel \mathcal{N}) \geq t$ indicates non-ideal parameters in the memristive NN. The specific choice of $t$ depends on several factors, which will be discussed later. 

Please note that other distance functions, such as the Jensen-Shannon divergence (which is a symmetrized version of the KL divergence), can also be used. Alternatively, for simplicity, evaluating only the $\mu_{\hat{\mathcal{N}}}$ and $\sigma_{\hat{\mathcal{N}}}$ values may be sufficient for fault and variation detection.

The test vector (stored in the hardware) can be applied periodically during online operation, and the deviation from the expected distribution can be used as an indicator for faults and variation in the memristive NN. The overall flow diagram of our one-shot testing approach is depicted in Fig.~\ref{fig:flow}.

\subsection{Proposed test vector generation method}

In order to make the proposed one-shot testing method possible, the distribution of $\hat{y}$ should not only be standardized but also done with a single test vector, i.e., \emph{one-shot}. We generate a special test vector for this purpose with a specific learning objective. However, there are several challenges associated with this.

\subsubsection{Learning objective}
Since our learning objective is to produce a standard Gaussian distribution for $\hat{y}$, several loss functions can be designed to encourage the $\hat{y}$ distribution $\hat{\mathcal{N}}$ to have a mean of $0$ and a standard deviation of $1$. For example:

\begin{equation}
    \argmin_{\mu_{\mathcal{N}}\rightarrow 0, \sigma_{\mathcal{N}}\rightarrow 1} \frac{1}{N} \sum_{i=1}^{N} \hat{y}_i \log \frac{\hat{y}_i}{y'_i},
\end{equation}

minimizes pointwise KL-divergence loss between NN output $\hat{y}$ and ground truth value $y'$. Alternatively,

\begin{equation}
    \argmin_{\mu_{\hat{\mathcal{N}}}\rightarrow 0, \sigma_{\hat{\mathcal{N}}}\rightarrow 1} (\mu_{\hat{\mathcal{N}}})^2 + (1-\sigma_{\hat{\mathcal{N}}})^2,
\end{equation}

encourages $\mu_{\hat{\mathcal{N}}}$ and $\sigma_{\hat{\mathcal{N}}}$ to be close to 0, and 1, respectively. Regression loss, such as

\begin{equation}
    \argmin_{\mu_{\hat{\mathcal{N}}}\rightarrow 0, \sigma_{\hat{\mathcal{N}}}\rightarrow 1} \frac{1}{N} \sum_{i=1}^{N} (\hat{y}_i-y'_i)^2,
\end{equation}
can also be used. Here, $N$ denotes the number of output classes in the NN.

The ground truth $y'$ for the training can be defined as

\begin{equation}
    y' = \frac{\hat{y}-\mu_{\hat{\mathcal{N} }}}{\sigma_{\hat{\mathcal{N}}}},
\end{equation}

or can be sampled from a unit Gaussian distribution. The number of samples should be the same as the number of output classes of a NN model. Our learning objective can be considered supervised learning.

The proposed one-shot test vector produces a standardized output distribution across different models and generated test vectors, as shown in the bottom half of Fig.~\ref{fig:distribution_change}(a) and (b). Additionally, the relative deviation of the output distribution for the one-shot test vector is significantly higher, as demonstrated in Fig.~\ref{fig:distribution_change}(c). As a result, our one-shot test vector is considerably more sensitive to non-ideal parameters.

\subsubsection{Initialization}
Let $\Bar{x}$ be the learnable one-shot testing vector with shape [H, W, C] (assuming a colored image) that is optimized based on the loss function~(\ref{eq:loss_KL}). Here, H, W, and C denote height, width, and number of channels, respectively. Initializing $\Bar{x}$ properly is crucial for proper learning, where initial values are assigned to each pixel of $\Bar{x}$ before training. The convergence speed and final loss value greatly depend on proper initialization. Additionally, appropriate initialization is essential for deeper networks, as the gradient is propagated all the way back to the input.

We initialize $\Bar{x}$ element-wise with random values drawn from a unit Gaussian distribution as follows:

\begin{equation}
    \Bar{x}_{i\rightarrow H,j\rightarrow W,k\rightarrow C} \sim \mathcal{N}(0, 1)
\end{equation}
Here, i, j, and k are the indices of the elements in the input tensor. Element-wise initialization enables fine-grained control over the initialization process and is commonly used in deep learning.

\begin{figure}
    \centering
     \includegraphics[width=\linewidth]{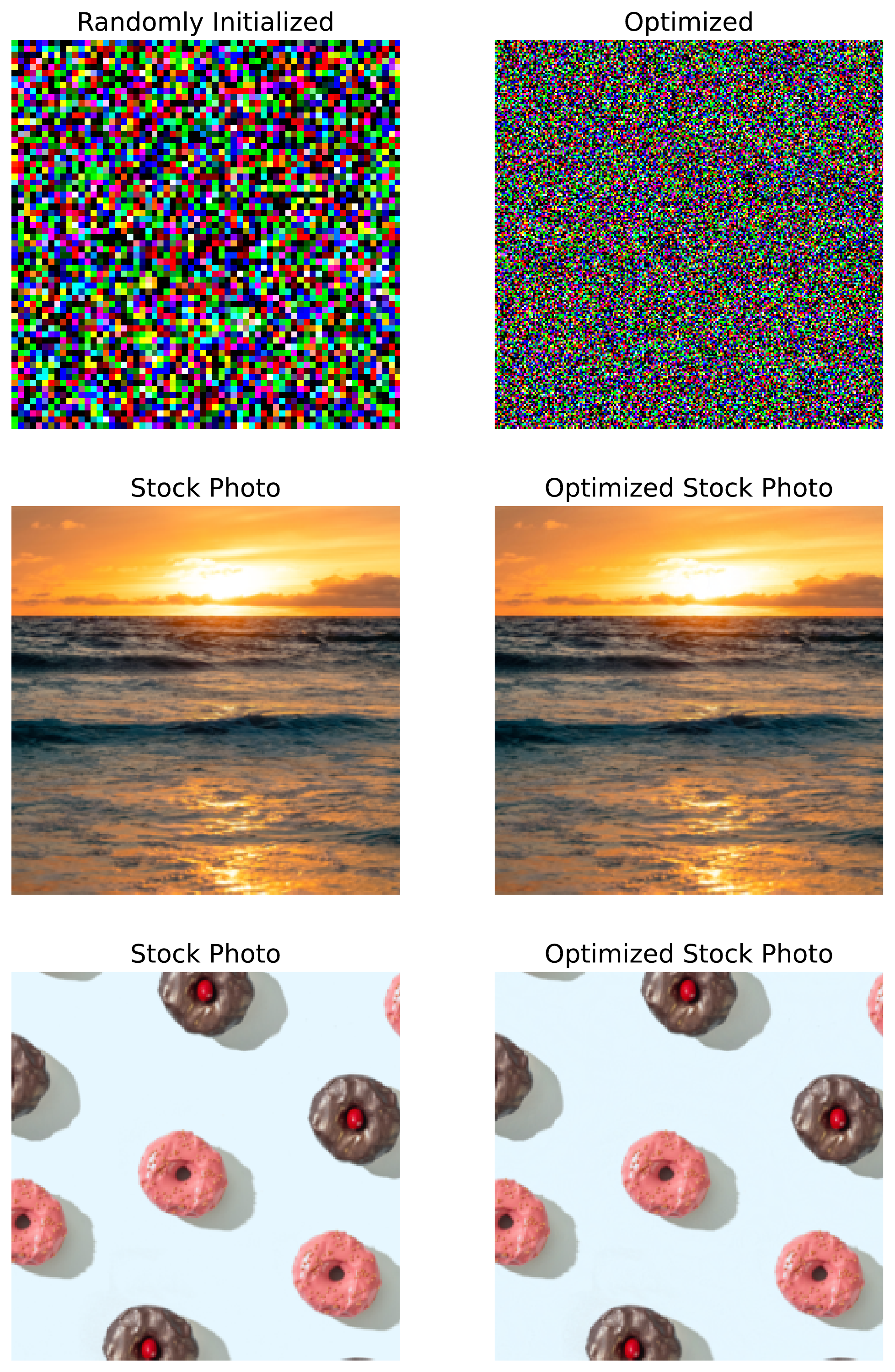}
    \caption{Some examples of the proposed one-shot test vector for the DenseNet-121 topology. To the naked eye, optimized stock images appear identical to their original images. Nevertheless, they differ marginally. }
    \label{fig:example_test_vect}
\end{figure}

Alternatively, initialization from out-of-distribution data, i.e. data that does not belong to the training set, also works well. This means that stock images from the internet can also be used for initialization. Therefore, access to training data is still not necessary. Fig.~\ref{fig:example_test_vect} shows some examples of generated test vectors with their initial images. Our optimization procedure makes minute adjustments to the stock photos. Therefore, they are visually indistinguishable.

The overall algorithm for the proposed one-shot test vector generation is summarized in Algorithm~\ref{alg:oneshot_decay}. To accelerate the learning process, we propose optimizing the test vector with an exponential decay learning rate for every $Q$ iteration. 



\begin{algorithm}
\caption{One-shot test vector generation using gradient descent with an exponential decaying learning rate}
\label{alg:oneshot_decay}
\begin{algorithmic}[1]
\REQUIRE 
\begin{tabular}[t]{@{}l}
Pre-trained network $\mathcal{F}$, loss function $\mathcal{L}(\hat{y}, y')$, initial \\learning rate $\alpha_0$, number of iterations $K$, Decay rate \\$Q$, and shape of the test vector $\Bar{x}$ [H, W, C].
\end{tabular}
\ENSURE One-shot test vector $\bar{x}$
\STATE Initialize $\bar{x}$ element-wise with random values from a \\unit Gaussian distribution
\FOR{$k = 1\ldots K$}
\STATE Perform forward pass through $\mathcal{F}$ with input $\bar{x}$ to obtain output $\hat{y}$
\STATE Compute loss $\mathcal{L}(\hat{y}, y')$
\STATE Calculate gradient $\nabla \mathcal{L}$ with respect to $\bar{x}$
\STATE Compute the current learning rate $\alpha_t$:
\begin{equation*}
\alpha_t = \begin{cases}
\alpha_0 & \text{if}\ t \bmod Q \neq 0 \\
\alpha_{t-1} / 10 & \text{if}\ t \bmod Q = 0
\end{cases}
\end{equation*}
\STATE Update $\bar{x}$ using gradient descent with the current learning rate: $\bar{x} \leftarrow \bar{x} - \alpha_t \nabla \mathcal{L}$
\ENDFOR
\end{algorithmic}
\end{algorithm}

\subsection{Relevance of Conventional Normalization Methods for Standardizing the Output Distribution}

As previously mentioned, normalization methods such as batch normalization standardize neuron activations before applying affine transformations. However, they are not suitable for our proposed one-shot testing method due to the following reasons:

\begin{enumerate}[label={\alph*)},font={\color{black!50!black}\bfseries}]
\item Conventional normalization methods are typically applied to intermediate activations (see Equation~\ref{eq:BN_eq}) of a NN and are not designed to directly standardize the output distribution, which is the primary goal of our one-shot testing method.

\item Batch normalization, as an example, requires multiple test vectors to estimate the mean and variance of the distribution, conflicting with the one-shot nature of our method that relies on a single test vector for output distribution standardization. Although other normalization techniques, such as group normalization~\cite{wu2018group}, have been proposed for small batch sizes, they are designed for specific tasks like sequence-to-sequence learning, recurrent neural networks (RNNs), or style transfer, and may not be directly applicable or easily adaptable to all deep learning tasks.

\item Finally, normalizing the model output may necessitate retraining the model using the entire training dataset, which could be computationally expensive, require access to the training data, and potentially negatively impact the model's performance, as it may not generalize well to unseen data.
\end{enumerate}

Thus, our unique approach to standardizing the output distribution aligns well with the requirements and objectives of our one-shot testing method.

\section{Simulation Results}\label{sec:result}
\subsection{Fault Modelling and Injection Framework}
\subsubsection{Modelling Conductance Variations}

Memristive technology and external environmental conditions influence conductance variation during online operations and in the manufacturing process. In this paper, we employ the variation model proposed in \cite{ahmed2022compact} which considers both device-to-device manufacturing variations (spatial fluctuation) and thermal variations (temporal fluctuation). This model injects multiplicative and additive Gaussian noise into the weight matrix of all layers as random noise, with a noise scale of $\eta_0$ used to control the severity of the noise. For each fault run, a different random sample is taken from the variation model.

\subsubsection{Modelling Online and Manufacturing Faults}
We consider two different kinds of fault models depending on the mapping employed: bit-wise and level-wise. As mentioned in Section~\ref{sec:background}, NN parameters can be encoded bit-wise, with eight memristive cells representing a single parameter. Our bit-wise fault model targets this kind of parameter encoding and can be expressed as: 

\begin{equation}\label{eq:permanent_noise}
    W_{flip} = f(\mathcal{P}_{flip}, W_{orig})
\end{equation}

Here, $\mathcal{P}_{flip}$ and $f(\cdot)$ represent the percentage of bit-flip faults injected and the fault model function, respectively. Specifically, the fault model $f(\cdot)$ randomly samples $\mathcal{P}_{flip}$\% of the bits of weights in each layer and flips their bits from $1$ to $0$ and vice versa. However, for parameter mapping with multi-level memristive cells, the (level-wise) fault model $f(\cdot)$ randomly sets the weights to a value between $-127$ and $127$. 

The ultimate effect of permanent faults is flipping the affected memristive cell from its desired level into another one. Therefore, the fault model $f(\cdot)$ considers read/write disturbance as well as permanent faults, including stuck-at faults.

\subsection{Simulation Setup}

In this paper, we have abstracted circuit-level details and evaluated our proposed one-shot approach using PyTorch-based simulation. We target  \emph{hard-to-detect} deviations in memristive crossbars. As the name suggests, detecting these kinds of deviations is hard, as they cause subtle changes in the output distribution and, in turn, inference accuracy. On the contrary, we have found that a large change in accuracy correlates to a large relative shift in the output distribution and is easier to detect with our approach in comparison.

Furthermore, instead of a simpler dataset like MNIST, we evaluated our method on larger pre-trained topologies, with up to 201 layers, trained on the more challenging ImageNet dataset~\cite{deng2009imagenet}, which is a large-scale image recognition dataset with 1000 classes, approximately 13 million training data points, and 50,000 validation data points. Additionally, we tested our approach on popular semantic segmentation topologies trained on both real-world brain MRI datasets and Microsoft's COCO benchmark dataset~\cite{buda2019association, lin2014microsoft}. Semantic segmentation is considerably more challenging than image classification, since it involves assigning labels to individual pixels in an image. Table~\ref{tab:evaluated_models} summarizes all the evaluated pre-trained models, their accuracy, and the number of parameters. All the pre-trained models are accessible through PyTorch Hub.

\renewcommand{\tabcolsep}{2pt}
\begin{table}[]
\caption{Showing the evaluated (pre-trained) models for classification and semantic segmentation tasks, along with their respective information such as inference accuracy, number of parameters, layers, and dataset used for training.}
\resizebox{\linewidth}{!}{
\begin{tabular}{|c|cccc|}
\hline
\multirow{2}{*}{Model} & \multicolumn{4}{c|}{Classification}                                                                                                                      \\ \cline{2-5} 
                       & \multicolumn{1}{c|}{Inference Acc.} & \multicolumn{1}{c|}{Parameters } & \multicolumn{1}{c|}{Layers} & Dataset                   \\ \hline\hline
ResNet-18~\cite{he2016deep}              & \multicolumn{1}{c|}{69.76\%}        & \multicolumn{1}{c|}{11.7$\times10^6$}                                & \multicolumn{1}{c|}{18}     & \multirow{6}{*}{Imagenet~\cite{deng2009imagenet}} \\ \cline{1-4}
ResNet-50~\cite{he2016deep}              & \multicolumn{1}{c|}{76.13\%}        & \multicolumn{1}{c|}{25.6$\times10^6$}                                & \multicolumn{1}{c|}{50}     &                           \\ \cline{1-4}
ResNet-101~\cite{he2016deep}             & \multicolumn{1}{c|}{81.89\%}        & \multicolumn{1}{c|}{44.5$\times10^6$}                                & \multicolumn{1}{c|}{101}    &                           \\ \cline{1-4}
DenseNet-121~\cite{huang2017densely}           & \multicolumn{1}{c|}{74.43\%}        & \multicolumn{1}{c|}{8$\times10^6$}                                   & \multicolumn{1}{c|}{121}    &                           \\ \cline{1-4}
DenseNet-201~\cite{huang2017densely}           & \multicolumn{1}{c|}{76.89\%}        & \multicolumn{1}{c|}{20$\times10^6$}                                  & \multicolumn{1}{c|}{201}    &                           \\ \cline{1-4}
MobileNet-V2~\cite{sandler2018mobilenetv2}          & \multicolumn{1}{c|}{71.87\%}        & \multicolumn{1}{c|}{3.5$\times10^6$}                                 & \multicolumn{1}{c|}{52}     &                           \\ \hline
                       & \multicolumn{4}{c|}{Semantic Segmentation}                                                                                                               \\ \hline
                       & \multicolumn{1}{c|}{Pixelwise Acc.} & \multicolumn{1}{c|}{Parameters} & \multicolumn{1}{c|}{Layers} & Dataset                   \\ \hline\hline
U-Net~\cite{ronneberger2015u}                   & \multicolumn{1}{c|}{98.75\%}              & \multicolumn{1}{c|}{7.76$\times10^6$}                                  & \multicolumn{1}{c|}{23}     & Brain MRI~\cite{buda2019association}                \\ \hline
DeepLab-V3~\cite{chen2017rethinking}             & \multicolumn{1}{c|}{91.2\%}         & \multicolumn{1}{c|}{11.03$\times10^6$}                                  & \multicolumn{1}{c|}{72}      & COCO~\cite{lin2014microsoft}                      \\ \hline
\end{tabular}
}
\label{tab:evaluated_models}
\end{table}
\renewcommand{\tabcolsep}{6pt}

For the fault coverage analysis, we have done the Monte Carlo simulation to simulate the effect of per-chip and online variations, as well as various faults modeled as bit-flip. Specifically, $1000$ memristive crossbars instances are evaluated for each noise level for variations and fault percentages. 

We report fault coverage as the ratio between detected faults ($D_{\mathrm{KL}}(\hat{\mathcal{N}} \parallel \mathcal{N}) \geq t$) and overall faults runs ($\mathcal{M}$) and can be described as 

\begin{equation}\label{eq:coverage}
    \text{fault coverage} = \frac{\text{\# of } D_{\mathrm{KL}}(\hat{\mathcal{N}} \parallel \mathcal{N}) \geq t }{\mathcal{M}}\times 100. 
\end{equation}

Note that, although we inject variation and faults into all parameters that are mapped to memristive cells, our fault coverage does not necessarily imply all possible faults that may occur.

\subsection{Detecting Variations in a One-Shot}

For classification tasks using the ImageNet dataset, Table~\ref{tab:coverage_var_imagenet} evaluates the fault coverage achieved by a proposed one-shot method on multiplicative and additive variations. The six state-of-the-art (SOTA) models consistently achieve $100\%$ fault coverage across various noise scales ($\eta_0$). Our result results indicate the robustness of the one-shot method in adapting to diverse levels of noise.

Similarly, for semantic segmentation tasks, as shown in Table~\ref{tab:coverage_var_segmentation}, the proposed one-shot method on multiplicative and additive variations with a range of noise scales ($\eta_0$) consistently achieves 100\% fault coverage. This further underscores the robustness of our one-shot method across different tasks.

\renewcommand{\tabcolsep}{1pt}
\begin{table*}[h!]
\centering
\caption{The fault coverage (\%) achieved by the proposed one-shot method on multiplicative and additive variations with different noise scales $\eta_0$. All the model is trained on the ImageNet dataset. }
\resizebox{\linewidth}{!}{
\begin{tabular}{|c|cccccc|ccccc|}
\hline
\multirow{2}{*}{Model} & \multicolumn{6}{c|}{Multiplicative Variations} & \multicolumn{5}{c|}{Additive Variations} \\ \cline{2-12}
                       & \multicolumn{1}{c|}{$\eta_0=0.01$} & \multicolumn{1}{c|}{$\eta_0=0.02$} & \multicolumn{1}{c|}{$\eta_0=0.04$} & \multicolumn{1}{c|}{$\eta_0=0.06$} & \multicolumn{1}{c|}{$\eta_0=0.08$} & $\eta_0=0.10$ & \multicolumn{1}{c|}{$\eta_0=0.0001$} & \multicolumn{1}{c|}{$\eta_0=0.0002$} & \multicolumn{1}{c|}{$\eta_0=0.00025$} & \multicolumn{1}{c|}{$\eta_0=0.0003$} & $\eta_0=0.00035$ \\ \hline\hline
ResNet-18              & \multicolumn{1}{c|}{100\%}     & \multicolumn{1}{c|}{100\%}     & \multicolumn{1}{c|}{100\%}     & \multicolumn{1}{c|}{100\%}     & \multicolumn{1}{c|}{100\%}     & 100\%     & \multicolumn{1}{c|}{100\%}       & \multicolumn{1}{c|}{100\%}       & \multicolumn{1}{c|}{100\%}        & \multicolumn{1}{c|}{100\%}       & 100\%        \\ \hline
ResNet-50              & \multicolumn{1}{c|}{100\%}     & \multicolumn{1}{c|}{100\%}     & \multicolumn{1}{c|}{100\%}     & \multicolumn{1}{c|}{100\%}     & \multicolumn{1}{c|}{100\%}     & 100\%     & \multicolumn{1}{c|}{100\%}       & \multicolumn{1}{c|}{100\%}       & \multicolumn{1}{c|}{100\%}        & \multicolumn{1}{c|}{100\%}       & 100\%        \\ \hline
ResNet-101             & \multicolumn{1}{c|}{100\%}     & \multicolumn{1}{c|}{100\%}     & \multicolumn{1}{c|}{100\%}     & \multicolumn{1}{c|}{100\%}     & \multicolumn{1}{c|}{100\%}     & 100\%     & \multicolumn{1}{c|}{100\%}       & \multicolumn{1}{c|}{100\%}       & \multicolumn{1}{c|}{100\%}        & \multicolumn{1}{c|}{100\%}       & 100\% \\ \hline
DenseNet-121           & \multicolumn{1}{c|}{100\%}     & \multicolumn{1}{c|}{100\%}     & \multicolumn{1}{c|}{100\%}     & \multicolumn{1}{c|}{100\%}     & \multicolumn{1}{c|}{100\%}     & 100\%     & \multicolumn{1}{c|}{100\%}       & \multicolumn{1}{c|}{100\%}       & \multicolumn{1}{c|}{100\%}        & \multicolumn{1}{c|}{100\%}       & 100\%        \\ \hline
DenseNet-201           & \multicolumn{1}{c|}{100\%}     & \multicolumn{1}{c|}{100\%}     & \multicolumn{1}{c|}{100\%}     & \multicolumn{1}{c|}{100\%}     & \multicolumn{1}{c|}{100\%}     & 100\%     & \multicolumn{1}{c|}{100\%}       & \multicolumn{1}{c|}{100\%}       & \multicolumn{1}{c|}{100\%}        & \multicolumn{1}{c|}{100\%}       & 100\%        \\ \hline
MobileNet-V2           & \multicolumn{1}{c|}{100\%}     & \multicolumn{1}{c|}{100\%}     & \multicolumn{1}{c|}{100\%}     & \multicolumn{1}{c|}{100\%}     & \multicolumn{1}{c|}{100\%}     & 100\%     & \multicolumn{1}{c|}{100\%}       & \multicolumn{1}{c|}{100\%}       & \multicolumn{1}{c|}{100\%}        & \multicolumn{1}{c|}{100\%}       & 100\%        \\ \hline
\end{tabular}
}
\label{tab:coverage_var_imagenet}
\end{table*}

\renewcommand{\tabcolsep}{6pt}
\begin{table*}[]
\caption{Fault coverage of semantic segmentation models utilizing the proposed one-shot testing method. The models are evaluated under multiplicative and additive variations with different noise scales $\eta_0$ to underscore the robustness of the models across different scenarios. }

\resizebox{\linewidth}{!}{
\begin{tabular}{|c|cccccc|}
\hline
\multirow{2}{*}{Model} & \multicolumn{6}{c|}{Additive Variations}                                                                                                                                                                                                                                                                     \\ \cline{2-7} 
                       & \multicolumn{1}{c|}{$\eta_0=0.00002$} & \multicolumn{1}{c|}{$\eta_0=0.00004$} & \multicolumn{1}{c|}{$\eta_0=0.00006$} & \multicolumn{1}{c|}{$\eta_0=0.00008$} & \multicolumn{1}{c|}{$\eta_0=0.00010$} & $\eta_0=0.00012$ \\ \hline\hline
U-Net                   & \multicolumn{1}{c|}{100\%}                          & \multicolumn{1}{c|}{100\%}                          & \multicolumn{1}{c|}{100\%}                          & \multicolumn{1}{c|}{100\%}                          & \multicolumn{1}{c|}{100\%}                          & 100\%                          \\ \hline
                       & \multicolumn{1}{c|}{$\eta_0=0.02$}    & \multicolumn{1}{c|}{$\eta_0=0.04$}    & \multicolumn{1}{c|}{$\eta_0=0.06$}    & \multicolumn{1}{c|}{$\eta_0=0.08$}    & \multicolumn{1}{c|}{$\eta_0=0.1$}     & $\eta_0=0.12$    \\ \hline
DeepLab-V3             & \multicolumn{1}{c|}{100\%}                          & \multicolumn{1}{c|}{100\%}                          & \multicolumn{1}{c|}{100\%}                          & \multicolumn{1}{c|}{100\%}                          & \multicolumn{1}{c|}{100\%}                          & 100\%                          \\ \hline
                       & \multicolumn{6}{c|}{Multiplicative Variations}                                                                                                                                                                                                                                                               \\ \hline
                       & \multicolumn{1}{c|}{$\eta_0=0.01$}    & \multicolumn{1}{c|}{$\eta_0=0.02$}    & \multicolumn{1}{c|}{$\eta_0=0.04$}    & \multicolumn{1}{c|}{$\eta_0=0.06$}    & \multicolumn{1}{c|}{$\eta_0=0.08$}    & $\eta_0=0.08$    \\ \hline\hline
U-Net                   & \multicolumn{1}{c|}{100\%}                          & \multicolumn{1}{c|}{99.9\%}                         & \multicolumn{1}{c|}{100\%}                          & \multicolumn{1}{c|}{100\%}                          & \multicolumn{1}{c|}{100\%}                          & 100\%                          \\ \hline
                       & \multicolumn{1}{c|}{$\eta_0=0.01$}    & \multicolumn{1}{c|}{$\eta_0=0.02$}    & \multicolumn{1}{c|}{$\eta_0=0.04$}    & \multicolumn{1}{c|}{$\eta_0=0.06$}    & \multicolumn{1}{c|}{$\eta_0=0.08$}    & $\eta_0=0.08$    \\ \hline
DeepLab-V3             & \multicolumn{1}{c|}{100\%}                          & \multicolumn{1}{c|}{100\%}                          & \multicolumn{1}{c|}{100\%}                          & \multicolumn{1}{c|}{100\%}                          & \multicolumn{1}{c|}{100\%}                          & 100\%                          \\ \hline
\end{tabular}
}
 \label{tab:coverage_var_segmentation}
\end{table*}

\subsection{Detecting Faults in a One-Shot}

For ImageNet classification on various SOTA topologies, Table \ref{tab:coverage_faults_classification} demonstrates a similarly high level of fault coverage under both bit-flip and level-flip fault conditions. For each model, as the fault rate increases, the percentage of fault coverage generally improves. At higher fault rates, such as 0.05\% and 0.1\%, most of the models achieved 100\% fault coverage. At lower fault rates, the shift in the output distribution is very low, resulting in a few false-negative cases. However, by reducing the threshold to a value closer to the KL-divergence value on a fault-free model, the number of false-negative cases can be reduced (see Table~\ref{tab:threshold}). Nevertheless, our results indicate the resilience of our one-shot approach to various types of faults at different rates.

\begin{table}[]
\caption{The effect of threshold $t$ on false-negative test cases.  As the threshold is reduced, the fault coverage increases.}
\centering
\begin{tabular}{|c|ccc|}
\hline
\multirow{2}{*}{Threshold} & \multicolumn{3}{c|}{\% of faults}                                         \\ \cline{2-4} 
                           & \multicolumn{1}{c|}{$\eta_0=0.02$} & \multicolumn{1}{c|}{$\eta_0=0.025$} & $\eta_0=0.33$   \\ \hline\hline
$1x10^{-4}$                     & \multicolumn{1}{c|}{99.7\%}    & \multicolumn{1}{c|}{99.2\%}     & 99.1\% \\ \hline
$1x10^{-5}$                     & \multicolumn{1}{c|}{99.4\%}    & \multicolumn{1}{c|}{99.8\%}     & 99.8\% \\ \hline
$1x10^{-6}$                     & \multicolumn{1}{c|}{99.9\%}    & \multicolumn{1}{c|}{99.8\%}     & 100\%  \\ \hline
$1x10^{-7}$                     & \multicolumn{1}{c|}{100\%}     & \multicolumn{1}{c|}{100\%}      & 100\%  \\ \hline
\end{tabular}
\label{tab:threshold}
\end{table}

Similarly, our proposed method can achieve a high fault-coverage on both bit-flip and level-flip fault conditions for semantic segmentation tasks on two state-of-the-art (SOTA) topologies, as demonstrated in Table~\ref{tab:coverage_faults_segmentation}. The trend in fault-coverage percentage for each model is similar to that of the models used for ImageNet classification. We also found that lowering the threshold can have a similar effect on fault-coverage, as observed in the ImageNet classification models.

\begin{table*}[]
\caption{Evaluation of the fault coverage of SOTA ImageNet classification models under different fault scenarios, including Bit-flip and Level-flip faults, and varying fault rates.}
\centering
\resizebox{\linewidth}{!}{
\begin{tabular}{|c|ccccc|ccccc|}
\hline
\multirow{3}{*}{Model} & \multicolumn{5}{c|}{Bit-flip}                                                                                                   & \multicolumn{5}{c|}{Level-flip}                                                                                                 \\ \cline{2-11} 
                       & \multicolumn{5}{c|}{\% of faults}                                                                                               & \multicolumn{5}{c|}{\% of faults}                                                                                               \\ \cline{2-11} 
                       & \multicolumn{1}{c|}{0.02\%} & \multicolumn{1}{c|}{0.025\%} & \multicolumn{1}{c|}{0.033\%} & \multicolumn{1}{c|}{0.05\%} & 0.1\% & \multicolumn{1}{c|}{0.02\%} & \multicolumn{1}{c|}{0.025\%} & \multicolumn{1}{c|}{0.033\%} & \multicolumn{1}{c|}{0.05\%} & 0.1\% \\ \hline\hline
ResNet-18              & \multicolumn{1}{c|}{98.6\%} & \multicolumn{1}{c|}{99.3\%}  & \multicolumn{1}{c|}{99.6\%}  & \multicolumn{1}{c|}{100\%}  & 100\% & \multicolumn{1}{c|}{99.9\%} & \multicolumn{1}{c|}{99.9\%}  & \multicolumn{1}{c|}{100\%}   & \multicolumn{1}{c|}{100\%}  & 100\% \\ \hline
ResNet-50              & \multicolumn{1}{c|}{99.6\%} & \multicolumn{1}{c|}{99.9\%}  & \multicolumn{1}{c|}{100\%}   & \multicolumn{1}{c|}{99.9\%} & 100\% & \multicolumn{1}{c|}{100\%}  & \multicolumn{1}{c|}{100\%}   & \multicolumn{1}{c|}{100\%}   & \multicolumn{1}{c|}{100\%}  & 100\% \\ \hline
ResNet-101             & \multicolumn{1}{c|}{98.9\%} & \multicolumn{1}{c|}{99.6\%}  & \multicolumn{1}{c|}{99.5\%}  & \multicolumn{1}{c|}{99.9\%} & 100\% & \multicolumn{1}{c|}{99.7\%} & \multicolumn{1}{c|}{100\%}   & \multicolumn{1}{c|}{100\%}   & \multicolumn{1}{c|}{100\%}  & 100\% \\ \hline
DenseNet-121           & \multicolumn{1}{c|}{99.9\%} & \multicolumn{1}{c|}{99.6\%}  & \multicolumn{1}{c|}{100\%}   & \multicolumn{1}{c|}{100\%}  & 100\% & \multicolumn{1}{c|}{99.9\%} & \multicolumn{1}{c|}{100\%}   & \multicolumn{1}{c|}{99.9\%}  & \multicolumn{1}{c|}{100\%}  & 100\% \\ \hline
DenseNet-201           & \multicolumn{1}{c|}{99.1\%} & \multicolumn{1}{c|}{99.7\%}  & \multicolumn{1}{c|}{99.8\%}  & \multicolumn{1}{c|}{100\%}  & 100\% & \multicolumn{1}{c|}{99.1\%} & \multicolumn{1}{c|}{100\%}   & \multicolumn{1}{c|}{100\%}   & \multicolumn{1}{c|}{100\%}  & 100\% \\ \hline
MobileNet-V2           & \multicolumn{1}{c|}{99.8\%} & \multicolumn{1}{c|}{99.8\%}  & \multicolumn{1}{c|}{100\%}   & \multicolumn{1}{c|}{100\%}  & 100\% & \multicolumn{1}{c|}{99.4\%} & \multicolumn{1}{c|}{99.8\%}  & \multicolumn{1}{c|}{99.8\%}  & \multicolumn{1}{c|}{100\%}  & 100\% \\ \hline
\end{tabular}
}
 \label{tab:coverage_faults_classification}
\end{table*}

\begin{table*}[]
\caption{Fault coverage performance of semantic segmentation models U-Net and DeepLab-V3 under Bit-flip and Level-flip fault scenarios with varying fault rates. 
}
\resizebox{\linewidth}{!}{
\begin{tabular}{|c|ccccc|ccccc|}
\hline
\multirow{3}{*}{Model} & \multicolumn{5}{c|}{Bit-flip}                                                                                                   & \multicolumn{5}{c|}{Level-flip}                                                                                                 \\ \cline{2-11} 
                       & \multicolumn{5}{c|}{\% of faults}                                                                                               & \multicolumn{5}{c|}{\% of faults}                                                                                               \\ \cline{2-11} 
                       & \multicolumn{1}{c|}{0.02\%} & \multicolumn{1}{c|}{0.025\%} & \multicolumn{1}{c|}{0.033\%} & \multicolumn{1}{c|}{0.05\%} & 0.1\% & \multicolumn{1}{c|}{0.02\%} & \multicolumn{1}{c|}{0.025\%} & \multicolumn{1}{c|}{0.033\%} & \multicolumn{1}{c|}{0.05\%} & 0.1\% \\ \hline\hline
U-Net                   & \multicolumn{1}{c|}{100\%}  & \multicolumn{1}{c|}{100\%}   & \multicolumn{1}{c|}{100\%}   & \multicolumn{1}{c|}{100\%}  & 100\% & \multicolumn{1}{c|}{99.9\%} & \multicolumn{1}{c|}{99.9\%}  & \multicolumn{1}{c|}{100\%}   & \multicolumn{1}{c|}{100\%}  & 100\% \\ \hline
DeepLab-V3             & \multicolumn{1}{c|}{100\%}  & \multicolumn{1}{c|}{100\%}   & \multicolumn{1}{c|}{100\%}   & \multicolumn{1}{c|}{100\%}  & 100\% & \multicolumn{1}{c|}{100\%}  & \multicolumn{1}{c|}{100\%}   & \multicolumn{1}{c|}{100\%}   & \multicolumn{1}{c|}{100\%}  & 100\% \\ \hline
\end{tabular}
}
 \label{tab:coverage_faults_segmentation}
\end{table*}

\subsection{Comparison with State of the Art and Overhead Analysis}

Our proposed method is compared against the related work that uses the functional test generation method and focuses on test pattern compaction. 
With only one test vector and test query, the proposed one-shot testing method outperforms existing methods \cite{chen2021line}, \cite{li2019rramedy}, \cite{luo2019functional}, and \cite{ahmed2022compact} on all metrics listed in Table~\ref{tab:comp_rel}. Therefore, the proposed method requires significantly fewer test vectors and queries compared to other methods. Furthermore, the proposed approach consistently achieves 100\% fault coverage, outperforming methods \cite{chen2021line, li2019rramedy, luo2019functional} which range from 76\% to 99.27\% coverage. Additionally, the proposed method is the most memory-efficient, requiring only 0.012288 MB, which is much lower than the other methods, regardless of whether re-training data is stored in hardware or not. Moreover, our method does not rely on storing re-training data in hardware to reduce memory consumption, unlike methods proposed by \cite{chen2021line}, and \cite{ahmed2022compact}.

The test application time (latency) and test energy are directly proportional to the number of test vectors used for testing. For example, the testing method \cite{li2019rramedy} requires $1024$ test vectors, therefore, their method requires $1024\times$ more matrix-vector operations and power consumption. In our comparisons, we assume the hardware implementation, NN topology, and NVM technology are the same. 

The analysis presented in Table~\ref{tab:comp_rel} is based on the numbers reported in related works. To calculate memory consumption, we utilized the bit-width reported in~\cite{chen2021line} for the images and test labels. Note that our approach does not require storing any labels.

\renewcommand{\tabcolsep}{4.5pt}
\begin{table}[h]
    \centering
    \footnotesize
    \caption{Compares the proposed approach with the existing methods using four performance metrics: fault coverage, memory storage overhead (with and without re-training data), number of test queries required, and fault-detection resolution. To ensure a fair comparison, the analysis of our approach is conducted on the CIFAR-10 dataset.
    }
    \begin{threeparttable}
\begin{tabular}{|c|c|c|c|c|c|}
    \hline 
    \multicolumn{1}{|c|}{} & \cite{chen2021line} &  \cite{li2019rramedy}&  \cite{luo2019functional}& \cite{ahmed2022compact} & \multicolumn{1}{c|}{ Proposed} \tabularnewline
    \hline 
    \multicolumn{1}{|c|}{Size of } & \multirow{2}{*}{10000} & \multirow{2}{*}{1024} & \multirow{2}{*}{${10}$-${50}$} & \multirow{2}{*}{$16$-$64$} & \multirow{2}{*}{$\bm{1}$} \tabularnewline
    \multicolumn{1}{|c|}{test vectors} &  &  &  & & \multicolumn{1}{c|}{} \tabularnewline
    \hline 
    \# of test queries  & \multirow{2}{*}{10000} & \multirow{2}{*}{1024} & \multirow{2}{*}{10-50} & \multirow{2}{*}{${17}$} & \multirow{2}{*}{$\bm{1}$} \tabularnewline
    (normalized)&  &  &  & & \multicolumn{1}{c|}{} \tabularnewline
    \hline 
    Memory & 0.015\tnote{1} & 234.42\tnote{1} & 0.154\tnote{1} & ${0.000256}$\tnote{1} & \multicolumn{1}{c|}{$\bm{0.012288}$\tnote{1}} \tabularnewline
    \cline{2-6}
    overhead (MB) & 245.775\tnote{2} & 234.42\tnote{2} & 0.154\tnote{2} & ${0.1966}$\tnote{2} & \multicolumn{1}{c|}{$\bm{0.012288}$\tnote{2}} \tabularnewline
    \hline 
    \multirow{1}{*}{Coverage (\%)} & 99.27\tnote{3} & 98\tnote{3} & 76\tnote{3} / 84\tnote{4} & ${100}$\tnote{4} & \multicolumn{1}{c|}{$\bm{100}$\tnote{3}} \tabularnewline
    \hline 
\end{tabular}
    
    \begin{tablenotes}
    \item[1] Re-training data is stored in hardware.
    \item[2] Re-training data is not stored in hardware.
    \item[3] Synthetic testing data
    \item[4] Original training data used as testing data
    \end{tablenotes}
    \end{threeparttable}
    
    \label{tab:comp_rel}
\end{table}

\subsection{Discussion and Future Works}

The proposed one-shot testing method determines the output distribution by calculating the mean and standard deviation. To avoid biased estimation of mean and standard deviation, it is important to have a sufficiently large number of output classes, such as 20 or more. For cases with fewer output classes like 2 or 5, alternative statistical methods like Bayesian approaches could be considered, but they are beyond the scope of this paper.

Additionally, the proposed one-shot testing method considers full precision as the bit-precision (32 bits floating point) of the test vector, allowing for high precision in the optimization process using the gradient descent algorithm. While quantizing the one-shot test vector can provide benefits such as reduced memory requirements and computational complexity, the limited representation ability and quantization error may impact the reliability of the output distribution estimations. As a result, a full-precision test vector is preferred.

\section{Conclusion}\label{sec:conclusion}
In this work, we have introduced a one-shot testing and respective test generation method to test the hardware accelerators for deep learning models based on memristor crossbars with a single test vector. Our method hypotheses that memristive non-idealities correlate to change in output distribution, and our testing method aims at detecting this distribution shift with a single testing vector. 
The proposed approach demonstrates superior performance in fault coverage, memory storage overhead, and the number of test queries required, highlighting its effectiveness and efficiency compared to the existing methods. Therefore, our work allows significantly faster detection of faults and variations at a negligible overhead.

\bibliographystyle{IEEEtran}
\typeout{}
\bibliography{references}


\end{document}